\documentclass[letterpaper, 10 pt, conference]{ieeeconf}
\IEEEoverridecommandlockouts

\usepackage{amsmath,amsfonts}
\usepackage{array}
\usepackage{textcomp}
\usepackage{graphicx}
\usepackage{booktabs}
\usepackage{placeins}
\usepackage{bm}
\usepackage{xcolor}
\usepackage{multirow}
\usepackage{float}
\usepackage{stfloats}
\usepackage[margin=0.75in, includefoot, letterpaper]{geometry}
\usepackage[colorlinks=true,linkcolor=blue,citecolor=blue,urlcolor=blue]{hyperref}
\usepackage{cite}
\usepackage{algorithm}
\let\labelindent\relax
\usepackage{algorithmic}
\usepackage{enumitem}

\title{An Open-Source Modular Benchmark for Diffusion-Based\\Motion Planning in Closed-Loop Autonomous Driving}

\author{Li Yun$^{1}$, Simon Thompson$^{2}$, Yidu Zhang$^{1}$, Ehsan Javanmardi$^{1}$, and Manabu Tsukada$^{1}$%
\thanks{$^{1}$Graduate School of Information Science and Technology, The University of Tokyo, Tokyo, Japan.
        {\tt\small \{li-yun, zhangyidu7, ejavanmardi\}@g.ecc.u-tokyo.ac.jp, tsukada@hongo.wide.ad.jp}}%
\thanks{$^{2}$TIER~IV, Inc., Tokyo, Japan. {\tt\small simon.thompson@tier4.jp}}%
}
\begin{document}
\maketitle
\thispagestyle{empty}

\begin{abstract}
Diffusion-based motion planners have achieved state-of-the-art results on benchmarks such as nuPlan, yet their evaluation within closed-loop production autonomous driving stacks remains largely unexplored.
Existing evaluations abstract away ROS\,2 communication latency and real-time scheduling constraints, while monolithic ONNX deployment freezes all solver parameters at export time.
We present an open-source modular benchmark that addresses both gaps: using ONNX GraphSurgeon, we decompose a monolithic 18{,}398-node diffusion planner into three independently executable modules and reimplement the DPM-Solver++ denoising loop in native C++.
Integrated as a ROS\,2 node within Autoware, the open-source AD stack deployed on real vehicles worldwide, the system enables runtime-configurable solver parameters without model recompilation and per-step observability of the denoising process, breaking the black box of monolithic deployment.
Unlike evaluations in standalone simulators such as CARLA, our benchmark operates within a production-grade stack and is validated through AWSIM closed-loop simulation.
Through systematic comparison of DPM-Solver++ (first- and second-order) and DDIM across six step-count configurations ($N \in \{3, 5, 7, 10, 15, 20\}$), we show that encoder caching yields a $3.2\times$ latency reduction, and that second-order solving reduces FDE by 41\% at $N{=}3$ compared to first-order.
The complete codebase will be released as open-source, providing a direct path from simulation benchmarks to real-vehicle deployment.
\end{abstract}

\begin{figure*}[!t]
\centering
\includegraphics[width=0.9\textwidth]{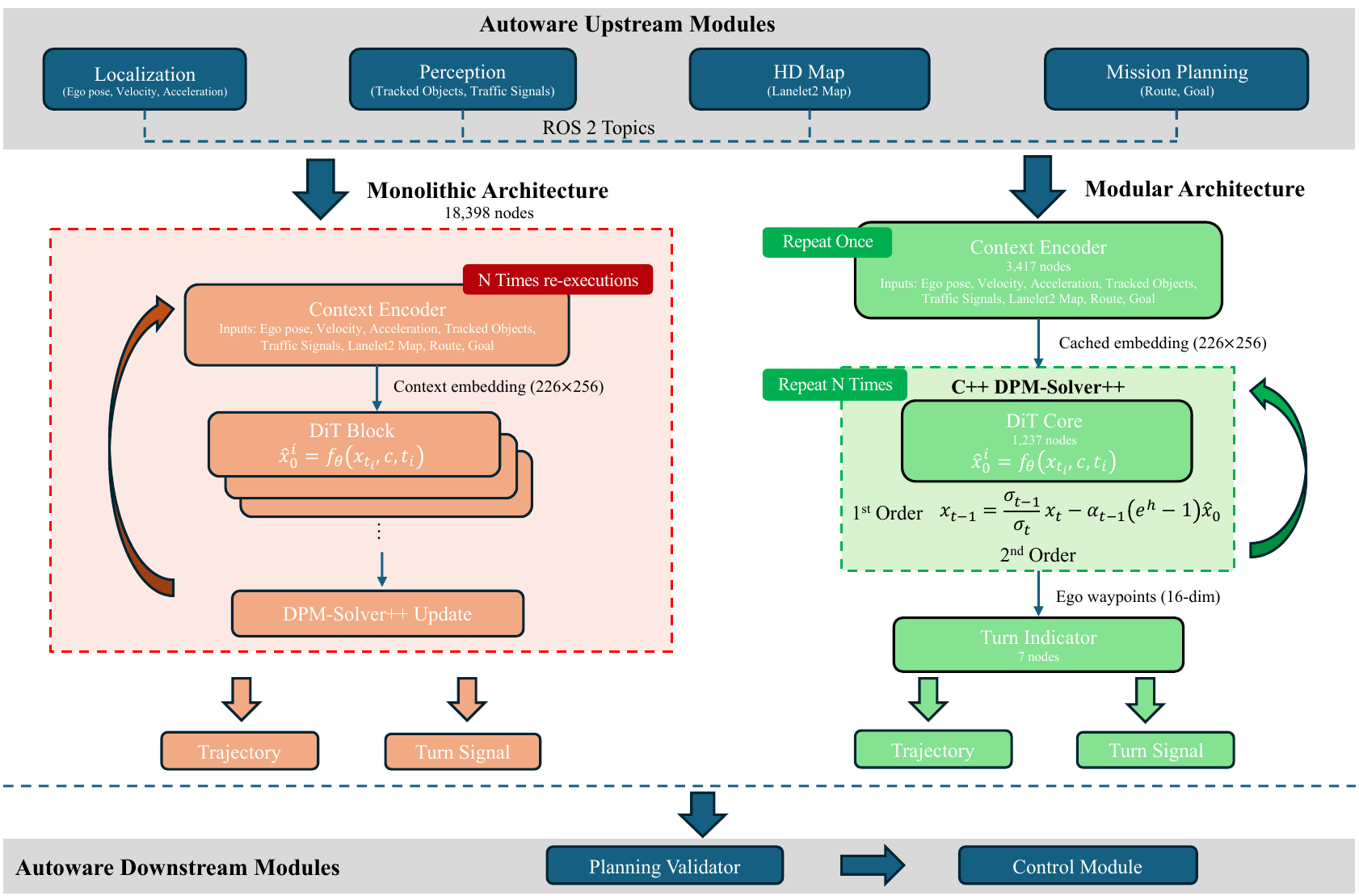}
\caption{\textbf{(a)}~Monolithic ONNX graph (18{,}398 nodes) compiled to TensorRT. \textbf{(b)}~Modular decomposition: three independently executable modules with the DPM-Solver++ loop in C++; the context encoder runs once and is cached across all denoising steps.}
\label{fig:framework}
\end{figure*}

\section{Introduction}
\label{sec:introduction}

Diffusion models have emerged as a powerful paradigm for motion planning in autonomous driving (AD)~\cite{Zheng2025-eq, Liao2024-js, Liu2025-uh, Zhang2025-eg}.
By formulating trajectory generation as iterative denoising from a learned data distribution, diffusion-based planners can jointly model multi-agent futures and produce temporally consistent plans.
Recent works such as Diffusion Planner~\cite{Zheng2025-eq}, DiffusionDrive~\cite{Liao2024-js}, and Latent Planner (LAP)~\cite{Zhang2025-eg} report state-of-the-art results on the nuPlan benchmark~\cite{Karnchanachari2024-ko, Dauner2023-eg}.
However, translating these benchmark results into deployable systems remains challenging, due to two largely unaddressed issues.

Nearly all existing diffusion planners are evaluated in \textit{pseudo-closed-loop} settings that replay recorded scenarios via Python-based simulators or nuPlan's reactive agents~\cite{Karnchanachari2024-ko}.
While useful for model comparison, these settings abstract away the latency constraints of production AD stacks: ROS\,2 topic communication overhead~\cite{Betz2023-zb, Betz2025-at}, sensor preprocessing pipelines, and real-time scheduling on shared vehicle compute.
Consequently, the gap between ``benchmark performance'' and ``vehicle-deployable performance'' remains unmeasured.
As shown in \S\ref{subsec:system_performance}, the monolithic diffusion planner studied in this work requires over 300\,ms per planning cycle on CPU, far exceeding the typical 100\,ms budget of Autoware-class planning nodes~\cite{Betz2023-zb}.

In addition, current deployment pipelines export the entire multi-step denoising process, including the noise schedule, solver logic, and iterative DiT blocks, into a single monolithic ONNX graph.
In the case of Diffusion Planner~\cite{Zheng2025-eq}, this produces 18{,}398 nodes, which are then compiled into a TensorRT engine.
This design freezes all solver parameters at export time: step count ($N{=}10$), solver order, and noise schedule cannot be changed without re-exporting and recompiling the model.
This prevents researchers from studying fundamental questions such as: \textit{How many denoising steps are truly needed under real-time constraints? Does second-order solving provide meaningful gains over first-order in closed-loop driving? Can anytime planning~\cite{Likhachev2003-tx} be realized by adaptively truncating diffusion steps?}

Both problems trace to a single design choice: embedding the iterative denoising loop \textit{inside} the computational graph.
Our key idea is to move the loop \textit{outside}, replacing the monolithic model with independently executable modules orchestrated by an external C++ solver.
Specifically, we use ONNX GraphSurgeon~\cite{nvidia_graphsurgeon} to decompose the monolithic graph into three modules:
a context encoder (3{,}417 nodes, 18\,MB), executed once per planning cycle and cached;
a DiT core (1{,}237 nodes, 28\,MB), invoked $N$ times where $N$ is configurable at runtime;
and a turn indicator classifier (7 nodes, 5.6\,KB) for post-hoc signal prediction.

The DPM-Solver++~\cite{Lu2022-nn} loop is reimplemented in native C++ with runtime-configurable step count ($N \in [1, 50]$), solver order, and VP noise schedule~\cite{Song2020-dn} parameters.
The system is integrated as a drop-in ROS\,2 node within the Autoware~\cite{autoware2025e2e} stack and validated in AWSIM~\cite{awsim2024}, the open-source simulator designed for closed-loop evaluation within the Autoware ecosystem.
The modular design creates a pluggable solver interface: researchers can substitute alternative denoising strategies (DDIM~\cite{Song2020-ip}, higher-order DPM-Solver variants, or novel schedulers) without modifying the neural network weights or retraining, and immediately evaluate them under realistic closed-loop conditions.

Our contributions are as follows.

First, we propose an open-source modular benchmark that decomposes a monolithic ONNX diffusion planner into independently executable modules via GraphSurgeon, enabling plug-and-play solver evaluation in the Autoware + AWSIM production AD stack under true closed-loop conditions.

Second, we provide a native C++ DPM-Solver++ implementation with runtime-configurable inference parameters, demonstrating that solver behavior, including step count, order, and noise schedule, can be studied without model recompilation, a capability absent from all existing diffusion planner deployments.

Third, through systematic solver comparison across DPM-Solver++ (first- and second-order) and DDIM over 50 AWSIM frames with $N \in \{3, 5, 7, 10, 15, 20\}$, we show that second-order correction reduces FDE by 41\% at $N{=}3$ and that proper timestep scheduling improves quality by 7$\times$ over naive early stopping.

Finally, we validate numerical equivalence with the monolithic model (max error $< 10^{-5}$), demonstrate encoder caching (90.9\% computation reduction, 3.2$\times$ latency improvement), and confirm successful urban driving in AWSIM.
The complete codebase will be released as open-source.

\section{Related Work}
\label{sec:related_work}

\subsection{Diffusion Models for Motion Planning}

Diffusion models~\cite{Ho2020-xt, Song2020-dn} have been increasingly applied to autonomous driving.
Diffuser~\cite{Janner2022-gh} pioneered diffusion-based trajectory planning for robotics;
Diffusion Policy~\cite{Chi2023-bj} demonstrated visuomotor control via action diffusion.
In the AD domain, Diffusion Planner~\cite{Zheng2025-eq} jointly models ego and neighbor trajectories using a DiT architecture~\cite{Peebles2023-vq} with a variance-preserving (VP) noise schedule, achieving state-of-the-art on the nuPlan benchmark~\cite{Karnchanachari2024-ko}.
DiffusionDrive~\cite{Liao2024-js} introduces truncated diffusion with anchored initialization, reducing steps to 2.
BridgeDrive~\cite{Liu2025-uh} formulates planning as a diffusion bridge; LAP~\cite{Zhang2025-eg} uses fine-grained latent features for fast inference; Diffusion-ES~\cite{Yang2024-dl} combines diffusion with gradient-free optimization; and DriveLaW~\cite{Xia2025-hx} unifies planning with video generation.
Despite these algorithmic advances, \textit{all} of these methods are evaluated on dataset replay or Python-based simulators, not within production AD stacks under real-time constraints.

\subsection{Closed-Loop Evaluation in Production Stacks}

The nuPlan benchmark~\cite{Karnchanachari2024-ko} provides a ``closed-loop'' evaluation by replaying scenarios with reactive agents.
However, this is a \textit{pseudo-closed-loop}: the planner runs in a Python process without the middleware, communication, and scheduling constraints present in production systems.
In real-world AD deployments, ROS\,2 serves as the standard middleware framework, providing the publish-subscribe communication infrastructure between perception, planning, and control modules.
Dauner~et~al.~\cite{Dauner2023-eg} showed that even simple rule-based planners can achieve competitive nuPlan scores, questioning whether benchmark rankings translate to real-world performance.
Betz~et~al.~\cite{Betz2023-zb} further demonstrated that ROS\,2-based AD stacks introduce significant communication latency that cascades through the full pipeline, fundamentally altering the operating conditions.
Our work bridges this gap by benchmarking within Autoware~\cite{autoware2025e2e}, a fully open-source ROS\,2-based AD stack with real-vehicle deployments by organizations worldwide.
Unlike proprietary platforms such as Baidu Apollo~\cite{apollo2024}, Autoware's open-source nature enables fully reproducible research with a direct path from simulation to on-vehicle deployment.

\subsection{Model Deployment and Partitioning}

Deploying DNN models on vehicle platforms typically involves ONNX export followed by hardware-specific compilation~\cite{onnxruntime2019, tensorrt2024}.
Guo~et~al.~\cite{Guo2024-zx} propose EASTER, which learns optimal split points for transformers across edge devices.
Nair~et~al.~\cite{Nair2023-ew} introduce SecureFrameNet for partitioning ONNX models in secure deployment scenarios.
Jajal~et~al.~\cite{Jajal2024-rt} analyze ONNX converter interoperability failures, identifying systematic issues across deep learning frameworks.
However, these approaches target general-purpose models and do not exploit the \textit{iterative} structure of multi-step diffusion models, where the same denoising subnetwork is invoked repeatedly within a solver loop.

\subsection{Diffusion Inference Acceleration}

A complementary line of work focuses on accelerating the diffusion sampling process itself.
DPM-Solver++~\cite{Lu2022-nn} is a high-order ODE solver for diffusion models that reduces the required denoising steps from hundreds as in DDPM~\cite{Ho2020-xt} to 10--20 while maintaining sample quality.
DDIM~\cite{Song2020-ip} enables deterministic sampling via an implicit non-Markovian process.
Step-reduction approaches such as truncated diffusion~\cite{Liao2024-js} reduce steps further but require model-specific training.
More recently, flow matching and consistency models have enabled one- or two-step generation, though their application to production AD planning remains limited.
Our work is \textit{orthogonal and complementary}. Rather than proposing a new solver, we provide an open-source platform where \textit{any} solver can be plugged in and evaluated under closed-loop driving conditions.

\section{System Architecture}
\label{sec:background}

\subsection{Diffusion Planner in the Autoware Stack}

Fig.~\ref{fig:framework} provides an overview of the system architecture.
Within the Autoware stack, the diffusion planner operates as the planning module, receiving perception outputs, including tracked objects, HD vector map, route, and traffic signals, and producing a trajectory that the downstream control module executes.
The Diffusion Planner~\cite{Zheng2025-eq} itself consists of three functional components:

\textbf{Context Encoder.}
The encoder processes the scene context $\mathcal{I} = \{s_\text{ego}, s_\text{nbr}, s_\text{lane}, s_\text{route}, s_\text{tl}, s_\text{goal}\}$, comprising ego vehicle history, neighbor trajectories, lane geometry, route information, traffic signals, and goal pose.
These heterogeneous inputs are fused through a mixer-attention architecture into a context embedding $\mathbf{c} \in \mathbb{R}^{L \times d}$ ($L{=}226$, $d{=}256$).

\textbf{Diffusion Transformer (DiT).}
The DiT backbone takes as input the noisy joint trajectory $\mathbf{x}_t \in \mathbb{R}^{A \times T \times 4}$ ($A{=}33$ agents, $T{=}81$ timesteps, state $(x, y, \cos\theta, \sin\theta)$), the context embedding $\mathbf{c}$, and a continuous timestep $t \in [0, 1]$.
It outputs a denoised trajectory estimate $\hat{\mathbf{x}}_0 \in \mathbb{R}^{A \times (T{-}1) \times 4}$.

\textbf{Turn Indicator Predictor.}
A linear classifier maps the context embedding $\mathbf{c}$ and sampled ego waypoints to a 4-class turn indicator prediction.

\subsection{VP Noise Schedule and DPM-Solver++}

Diffusion models~\cite{Ho2020-xt, Song2020-dn} generate data by learning to reverse a noise-adding forward process.
Given a clean trajectory $\mathbf{x}_0$, the forward process produces a noisy version $\mathbf{x}_t = \alpha(t)\mathbf{x}_0 + \sigma(t)\boldsymbol{\epsilon}$ at continuous time $t \in [0, 1]$, where $\boldsymbol{\epsilon} \sim \mathcal{N}(\mathbf{0}, \mathbf{I})$.
The VP noise schedule~\cite{Song2020-dn} parameterizes the signal and noise magnitudes as:
\begin{equation}
\log \alpha(t) = -\tfrac{1}{4}t^2(\beta_1 - \beta_0) - \tfrac{1}{2}t\beta_0,
\end{equation}
with $\sigma(t) = \sqrt{1 - \alpha(t)^2}$.
The log signal-to-noise ratio $\lambda(t) = \log[\alpha(t)/\sigma(t)]$ decreases monotonically from clean data ($t{\approx}0$) to near-pure noise ($t{\approx}1$).
The Diffusion Planner uses $\beta_0 = 0.1$ and $\beta_1 = 20.0$.

At inference, the DiT predicts the clean trajectory $\hat{\mathbf{x}}_0$ from the noisy input $\mathbf{x}_t$ and context $\mathbf{c}$, and a solver iteratively denoises from $t{=}1$ toward $t{=}0$.
DPM-Solver++~\cite{Lu2022-nn} is a dedicated high-order ODE solver for this reverse process.
The first-order update from timestep $t_s$ to $t$ is:
\begin{equation}
\mathbf{x}_t = \frac{\sigma_t}{\sigma_s}\mathbf{x}_s + \alpha_t(1 - e^{-h})\hat{\mathbf{x}}_0,
\label{eq:first_order}
\end{equation}
where $h = \lambda_t - \lambda_s$ is the logSNR difference between the two timesteps.
When a previous prediction is available, the second-order update refines the estimate by incorporating a finite-difference correction:
\begin{equation}
\mathbf{x}_t = \frac{\sigma_t}{\sigma_s}\mathbf{x}_s + \alpha_t(1 - e^{-h})\hat{\mathbf{x}}_0 + \frac{1}{2}\alpha_t(1 - e^{-h})D_1,
\label{eq:second_order}
\end{equation}
where $D_1 = (\hat{\mathbf{x}}_0^{(i)} - \hat{\mathbf{x}}_0^{(i-1)}) / r$ is the finite-difference derivative with $r = h_{i-1}/h_i$.
This correction leverages consecutive predictions to better approximate the denoising trajectory, achieving higher accuracy with the same number of neural network evaluations.

\subsection{Monolithic Deployment}
\label{subsec:monolithic}

In the original Autoware integration, the entire diffusion process, including 11 DiT steps, the VP schedule, and the DPM-Solver++ logic, is \textit{unrolled} into a single ONNX graph at export time.
This produces a model with \textbf{18{,}398 nodes}, compared to 1{,}237 for a single DiT step, that is compiled into a $\sim$68\,MB TensorRT engine.
While this enables aggressive kernel fusion, it freezes all solver parameters and forces the context encoder to re-execute at every denoising step.
The total inference cost per planning cycle is:
\begin{equation}
T_\text{mono} = (N{+}1) \cdot T_\text{enc} + (N{+}1) \cdot T_\text{dit} + N \cdot T_\text{sol},
\label{eq:latency_mono}
\end{equation}
where $T_\text{enc}$, $T_\text{dit}$, and $T_\text{sol}$ denote the latency of the context encoder, a single DiT step, and the solver update, respectively.
Since $T_\text{enc} \gg T_\text{dit}$ as quantified in \S\ref{subsec:system_performance}, the redundant $(N{+}1)$ encoder invocations dominate the total cost.
Fig.~\ref{fig:framework}(a) illustrates this monolithic design; our proposed modular alternative is detailed in \S\ref{sec:methodology} and shown in Fig.~\ref{fig:framework}(b).

\section{Methodology}
\label{sec:methodology}

\subsection{Graph Decomposition via ONNX GraphSurgeon}
\label{subsec:decomposition}

As shown in Fig.~\ref{fig:framework}(a), the monolithic ONNX graph contains a \textit{repeated substructure}: the same DiT block appears 11 times with shared weights but different timestep inputs.
We exploit this structure to decompose the graph into three independently executable modules (Fig.~\ref{fig:framework}(b)).

\textbf{Step 1: Subgraph Identification.}
Using ONNX GraphSurgeon~\cite{nvidia_graphsurgeon}, we analyze the computational graph to identify:
(a)~the context encoder subgraph, whose output feeds into all DiT blocks but receives no feedback from them;
(b)~a single DiT block, identified by weight sharing across the 11 unrolled copies;
and (c)~the turn indicator linear layer at the graph's output.

\textbf{Step 2: Subgraph Extraction.}
Each subgraph is extracted into an independent ONNX model with explicitly defined inputs and outputs.
The encoder's output node (context embedding $\mathbf{c}$) becomes both the encoder model's output and the DiT model's input.
The timestep, which was a constant in each unrolled copy, becomes a dynamic input to the extracted DiT model.
Concretely, we use GraphSurgeon to create new input \texttt{Variable} nodes at each subgraph boundary, specifically the encoder's final LayerNormalization output and the DiT's output projection, rewire all downstream consumer nodes, and invoke automatic graph cleanup to prune the now-unreachable encoder and duplicate DiT nodes.
The turn indicator is reconstructed as a minimal 7-node graph from the extracted linear layer weights.

\textbf{Step 3: Validation.}
We verify numerical equivalence by running both the original monolithic model and the decomposed pipeline on identical inputs, confirming a maximum element-wise error below $10^{-5}$; see~\S\ref{subsec:numerical}.

Table~\ref{tab:model_comparison} summarizes the decomposition.

\begin{table}[t]
\centering
\caption{Monolithic vs.\ modular architecture.}
\label{tab:model_comparison}
\footnotesize
\begin{tabular}{lcc}
\toprule
& \textbf{Monolithic} & \textbf{Modular (ours)} \\
\midrule
ONNX nodes & 18{,}398 & 4{,}661$^\dagger$ \\
Model size & 47\,MB & 46\,MB \\
Enc.\ calls / cycle & $N{+}1$ & 1 (cached) \\
Runtime backend & TensorRT (GPU) & ONNX Runtime (CPU/GPU) \\
Solver config & Frozen at export & Runtime-configurable \\
\bottomrule
\multicolumn{3}{l}{\footnotesize $^\dagger$Encoder (3{,}417) + DiT core (1{,}237) + Turn indicator (7).} \\
\end{tabular}
\end{table}

Unlike the monolithic TensorRT engine, the decomposed ONNX modules can be executed via ONNX Runtime~\cite{onnxruntime2019} on any supported backend, including CPU, CUDA, and other accelerators, without recompilation.

\subsection{C++ DPM-Solver++ Implementation}
\label{subsec:solver}

With the DiT core extracted as a single-step model, we reimplement the DPM-Solver++ denoising loop in native C++.
This choice is driven by the Autoware ecosystem: Autoware nodes are implemented in C++, the ONNX Runtime C++ API avoids Python interpreter overhead and GIL constraints, and deterministic memory management is essential for meeting the real-time scheduling requirements of production AD stacks.
Algorithm~1 presents the full procedure.

\begin{algorithm}[t]
\caption{Modular Diffusion Planning with DPM-Solver++}
\label{alg:dpm_solver}
\small
\begin{algorithmic}[1]
\REQUIRE Scene inputs $\mathcal{I}$, steps $N$, order $p \in \{1, 2\}$
\STATE $\mathbf{c} \leftarrow \text{Encoder}(\mathcal{I})$ \hfill $\triangleright$ \textit{run once, cache}
\STATE $\{t_0, \ldots, t_N\} \leftarrow \text{VPSchedule}(N, \beta_0, \beta_1)$
\STATE $\mathbf{x}_{t_0} \leftarrow \mathbf{0}$ \hfill $\triangleright$ \textit{zero init (temp.$= 0$)}
\FOR{$i = 0$ \TO $N-1$}
\STATE $\hat{\mathbf{x}}_0^{(i)} \leftarrow \text{DiT}(\mathbf{x}_{t_i}, \mathbf{c}, t_i)$ \hfill $\triangleright$ \textit{single step}
\IF{$i = 0$ \OR $p = 1$}
\STATE $\mathbf{x}_{t_{i+1}} \leftarrow$ Eq.~(\ref{eq:first_order}) \hfill $\triangleright$ \textit{1st-order}
\ELSE
\STATE $\mathbf{x}_{t_{i+1}} \leftarrow$ Eq.~(\ref{eq:second_order}) \hfill $\triangleright$ \textit{2nd-order}
\ENDIF
\STATE $\mathbf{x}_{t_{i+1}}[\text{ego}, 0] \leftarrow \mathbf{s}_{\text{cur}}$ \hfill $\triangleright$ \textit{constrain}
\ENDFOR
\STATE $\hat{\mathbf{x}}_0 \leftarrow \text{DiT}(\mathbf{x}_{t_N}, \mathbf{c}, t_N)$ \hfill $\triangleright$ \textit{denoise-to-zero}
\STATE $\mathbf{y}_\text{turn} \leftarrow \text{TurnInd}(\hat{\mathbf{x}}_0, \mathbf{c})$
\STATE $\hat{\mathbf{x}}_0 \leftarrow \text{Denorm}(\hat{\mathbf{x}}_0)$
\RETURN $\hat{\mathbf{x}}_0, \mathbf{y}_\text{turn}$
\end{algorithmic}
\end{algorithm}

The timestep schedule follows logSNR-uniform spacing~\cite{Lu2022-nn}: given $N$ steps, we uniformly partition the logSNR range $[\lambda_{\min}, \lambda_{\max}]$ into $N{+}1$ values and invert each to obtain continuous timesteps $t_i \in (0, 1]$.
This concentrates steps in the high-noise region where the denoising trajectory changes most rapidly.
After each solver update, the current-timestep slice of the trajectory tensor is overwritten with observed ego and neighbor agent states, anchoring the prediction to physical reality, as shown in Algorithm~1, line~9.

A key benefit of the modular design is encoder caching.
Because the context encoder output $\mathbf{c}$ depends only on the scene context $\mathcal{I}$ and not on the denoising state $\mathbf{x}_t$, it can be computed once and reused for all $N{+}1$ DiT invocations, reducing the inference cost from $T_\text{mono}$ in Eq.~\ref{eq:latency_mono} to:
\begin{equation}
T_\text{mod} = T_\text{enc} + (N{+}1) \cdot T_\text{dit} + N \cdot T_\text{sol}.
\label{eq:latency_mod}
\end{equation}
The saving of $N \cdot T_\text{enc}$ is substantial because $T_\text{enc} \gg T_\text{dit}$ (\S\ref{subsec:system_performance}).
The model operates in a normalized coordinate frame; output trajectories are denormalized to world coordinates before being published as ROS\,2 trajectory messages, ready for direct consumption by downstream planning validation and vehicle control modules within the Autoware stack.

\subsection{ROS\,2 Integration with Autoware}
\label{subsec:integration}

The modular planner is implemented as a drop-in replacement for the original Autoware diffusion planner node.
The ROS\,2 node subscribes to odometry, tracked objects, vector map, route, and traffic signal topics, performs preprocessing identical to the original node, and publishes trajectory and turn indicator outputs.
All solver parameters ($N$, $p$, $\beta_0$, $\beta_1$) are exposed as ROS\,2 parameters, enabling runtime reconfiguration via \texttt{ros2 param set} without node restart.

\section{Experiments}
\label{sec:experiments}

\subsection{Experimental Setup}

\begin{figure}[t]
\centering
\includegraphics[width=\columnwidth]{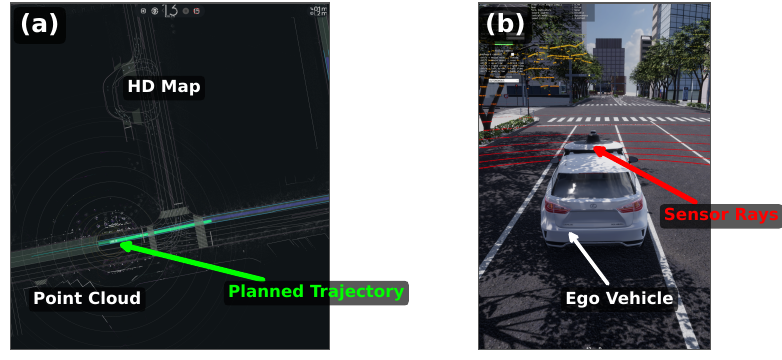}
\caption{Closed-loop evaluation environment. \textbf{(a)}~RViz visualization showing the HD map, point cloud, and planned trajectory within the Autoware stack. \textbf{(b)}~AWSIM 3D simulation at an urban intersection in the Nishishinjuku map, with the ego vehicle and sensor perception rays.}
\label{fig:simulation}
\end{figure}

All experiments are conducted on a desktop PC with an Intel Core i9-12900KS CPU, 32\,GB RAM, and an NVIDIA GeForce RTX 3090\,Ti GPU, running in a Docker container with Ubuntu 22.04, ROS\,2 Humble, and Autoware.
The modular planner uses ONNX Runtime 1.23 with the CPU execution provider; the original monolithic planner~\cite{tieriv_diffusion_planner} uses TensorRT with the 68\,MB engine file on GPU.
Both planners are benchmarked on the same machine; the CPU-based comparison isolates the architectural benefit of encoder caching from hardware acceleration effects.
Closed-loop evaluation is performed in AWSIM~\cite{awsim2024}, the open-source simulator integrated with Autoware~\cite{autoware2025e2e}, on an urban driving route in the Nishishinjuku map involving straight roads, intersections, and traffic participants.
Unless stated otherwise, the modular planner uses $N=10$ steps, second-order solver ($p=2$), VP schedule with $\beta_0=0.1$, $\beta_1=20.0$, and temperature $=0$ for deterministic sampling with zero initial noise, consistent with the original deployment.
Fig.~\ref{fig:simulation} shows the evaluation environment: the RViz visualization of the Autoware stack (left) and the AWSIM 3D simulation (right).

To verify that graph decomposition preserves model semantics\label{subsec:numerical}, we compare outputs of the original monolithic model~\cite{tieriv_diffusion_planner} and the modular pipeline on captured AWSIM driving data.
For each output tensor $\mathbf{y}$, we measure the maximum element-wise absolute error:
\begin{equation}
\epsilon = \max_{i}\left|y_i^{\text{mono}} - y_i^{\text{mod}}\right|.
\label{eq:numerical_error}
\end{equation}
For individual modules such as the context encoder and a single DiT step, $\epsilon < 10^{-5}$.
After 10 sequential solver steps, accumulated floating-point rounding yields $\epsilon < 10^{-4}$ for the final trajectory.
Since the planner operates in a normalized coordinate frame with standard deviation $\sigma = 20$\,m, the denormalization $x_{\text{phys}} = x_{\text{norm}} \cdot \sigma + \mu$ bounds the physical-space error to $\epsilon \cdot \sigma < 10^{-4} \times 20 = 2 \times 10^{-3}$\,m, i.e., less than 2\,mm, negligible for planning.

\subsection{Denoising Analysis and Solver Comparison}
\label{subsec:solver_study}

\begin{figure}[t]
\centering
\includegraphics[width=\columnwidth]{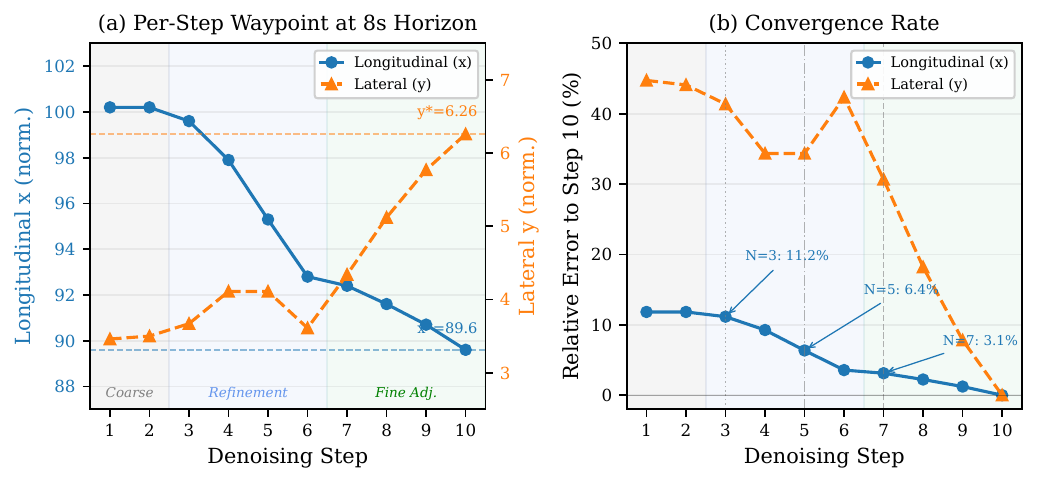}
\caption{Denoising progression of the 8s-horizon waypoint. \textbf{(a)}~Per-step predicted position in normalized coordinates: longitudinal (\textcolor{blue}{blue}) and lateral (\textcolor{orange}{orange}); three phases are visible: coarse (steps 1--2), refinement (3--6), and fine adjustment (7--10). \textbf{(b)}~Relative error to the converged step-10 prediction.}
\label{fig:denoising_convergence}
\end{figure}

A key capability of the modular architecture is \textit{per-step observability}: because the C++ solver explicitly calls the DiT core at each step, we can log intermediate predictions, information that remains a black box in the monolithic deployment.
Fig.~\ref{fig:denoising_convergence}(a) illustrates this using actual data from an AWSIM driving frame.
The denoising process exhibits three distinct phases: a \textit{coarse} phase (steps 1--2) where the prediction remains near the noise prior with minimal change, a \textit{refinement} phase (steps 3--6) where the largest per-step changes occur as both axes converge rapidly, and a \textit{fine adjustment} phase (steps 7--10) where residual corrections diminish.
The lateral position refines non-monotonically, indicating that curvature adjustment lags behind the longitudinal structure.
Fig.~\ref{fig:denoising_convergence}(b) quantifies the convergence rate: the longitudinal relative error drops to 6.4\% by step 5 and below 3.1\% by step 7, while the lateral channel converges more slowly due to its non-monotonic refinement.
This diagnostic capability, combined with runtime-configurable $N$, $p$, and $(\beta_0, \beta_1)$, enables systematic solver analysis without model recompilation.

Since the refinement phase concentrates most of the prediction change, reducing step count primarily truncates the fine adjustment phase.
At $N{=}7$, with 27\% fewer DiT calls, the mean 8s-horizon waypoint deviation is only 0.87\,m, well within the replanning margin of a 10\,Hz planner that executes only the first 1--2\,s of each trajectory.
Even at $N{=}5$ with 45\% reduction, the 1.71\,m deviation may be acceptable for latency-critical scenarios, supporting the feasibility of anytime planning~\cite{Likhachev2003-tx} where step count adapts to available compute budget.

\begin{figure*}[!t]
\centering
\includegraphics[width=0.75\textwidth]{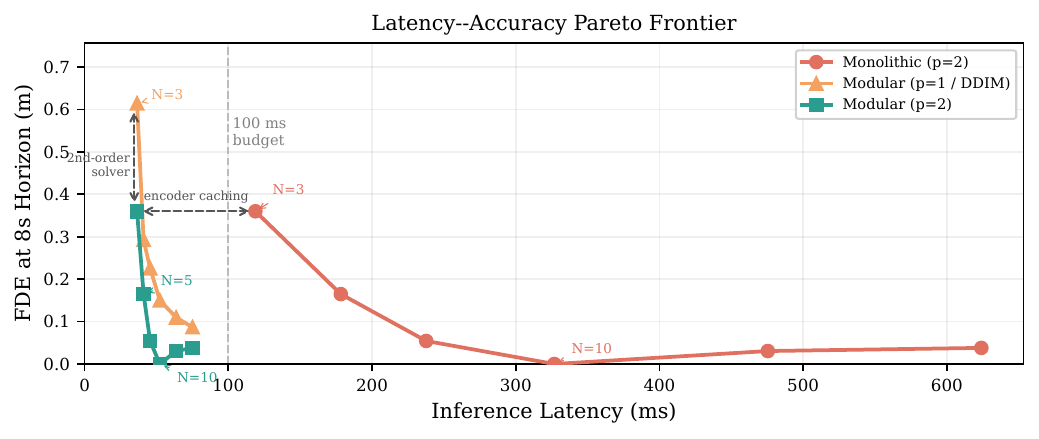}
\caption{Latency--accuracy Pareto frontier. Horizontal gap: encoder caching benefit; vertical gap: solver order benefit. Dashed line: 100\,ms planning budget.}
\label{fig:pareto}
\end{figure*}

\begin{figure}[t]
\centering
\includegraphics[width=\columnwidth]{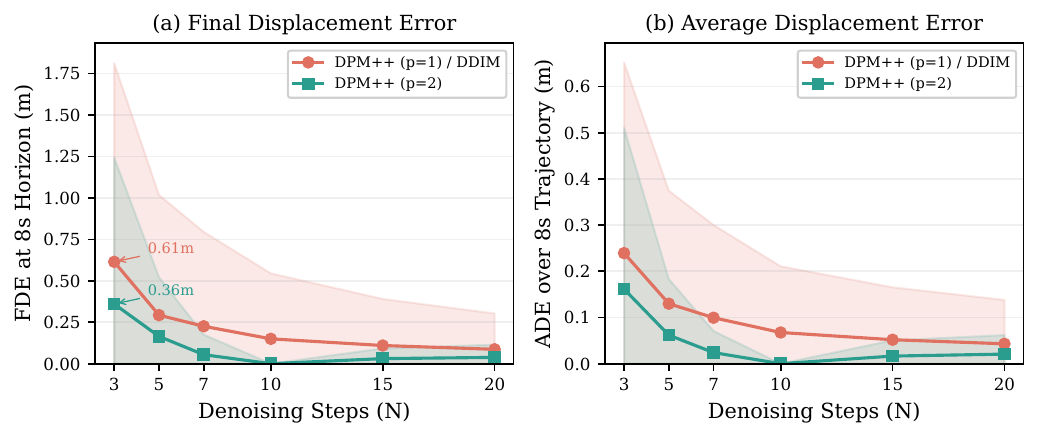}
\caption{Solver comparison. \textbf{(a)}~FDE and \textbf{(b)}~ADE vs.\ $N{=}10$, $p{=}2$ reference. Shaded: $\pm 1\sigma$.}
\label{fig:solver_comparison}
\end{figure}

To further demonstrate the pluggable solver interface, we conduct an offline benchmark: for each of 50 evenly sampled AWSIM frames, we replay the captured input tensors through the DiT core with three solver configurations (DPM-Solver++ first-order ($p{=}1$), second-order ($p{=}2$), and DDIM~\cite{Song2020-ip}) across $N \in \{3, 5, 7, 10, 15, 20\}$.
Each configuration uses its own logSNR-uniform timestep schedule optimized for the given $N$.
Fig.~\ref{fig:solver_comparison} reports Final Displacement Error (FDE) and Average Displacement Error (ADE) relative to the $N{=}10$, $p{=}2$ reference.
DPM-Solver++ first-order produces outputs identical to DDIM across all configurations, empirically confirming the known theoretical equivalence for data-prediction models~\cite{Lu2022-nn}.
The second-order solver provides substantial gains. At $N{=}3$, FDE drops by 41\% relative to first-order, from 0.61\,m to 0.36\,m, and the improvement is consistent across all $N$, with $p{=}2$ achieving the same accuracy as $p{=}1$ at roughly half the steps.
Proper timestep scheduling is equally important: comparing with the early-stopping analysis above, a dedicated $N{=}3$ schedule yields 0.36\,m FDE vs.\ 2.53\,m when truncating a 10-step schedule, a 7$\times$ improvement.

Because the modular architecture exposes the solver as a configurable component, researchers can directly evaluate novel scheduling strategies, step-adaptive policies, and alternative solvers on driving data captured from a production AD stack, rather than relying on dataset replay in standalone simulators.

\subsection{System Performance and Encoder Caching}
\label{subsec:system_performance}

While the previous sections analyzed algorithmic convergence, this section evaluates the system-level metric that matters for autonomous driving: \textit{inference latency}.

Table~\ref{tab:latency_breakdown} decomposes the inference latency, measured via ONNX Runtime on CPU over 100 runs with the median reported.
The context encoder dominates the per-step cost at $T_{\text{enc}} = 27.5$\,ms.
In the monolithic architecture, this encoder is redundantly re-executed at every denoising step due to the fused graph structure, a 90.9\% computational waste.
Our modular architecture runs it \textit{once} and caches the output, reducing the marginal cost of each additional step to just $T_{\text{dit}} = 2.3$\,ms.
Note that the monolithic 328\,ms in Table~\ref{tab:latency_breakdown} is the equivalent CPU cost computed from the same per-component latencies; the actual monolithic planner runs on GPU via TensorRT, so this value isolates the architectural difference rather than the hardware difference.

\begin{table}[t]
\centering
\caption{Inference latency breakdown.}
\label{tab:latency_breakdown}
\footnotesize
\begin{tabular}{lccc}
\toprule
\textbf{Component} & \textbf{Time} & \textbf{Mono.\ calls} & \textbf{Mod.\ calls} \\
\midrule
Context Encoder & 27.5\,ms & $N{+}1$ & 1 \\
DiT Core$^\dagger$ (per step) & 2.3\,ms & $N{+}1$ & $N{+}1$ \\
C++ Solver overhead & 0.01\,ms/step & --- & $N$ \\
\midrule
\textbf{Total ($N{=}10$)} & & \textbf{328\,ms} & \textbf{53\,ms} \\
\bottomrule
\multicolumn{4}{l}{\footnotesize $^\dagger$$N{+}1$: $N$ solver steps + 1 final denoise-to-zero (Alg.~1, line 10).} \\
\end{tabular}
\end{table}

Fig.~\ref{fig:pareto} plots inference latency against FDE for three configurations; the ideal operating point lies in the lower-left corner, where both latency and prediction error are minimized.
The horizontal arrow in the figure illustrates the encoder caching benefit: at $N{=}3$ with the same second-order solver, switching from monolithic to modular reduces latency from 119\,ms to 37\,ms, a \textbf{3.2$\times$} speedup at identical FDE.
The vertical arrow shows the solver order benefit: at the same latency, second-order solving at $N{=}3$ matches the accuracy of first-order at $N{=}5$, effectively halving the required steps.
Notably, \textit{all} modular configurations fall within the 100\,ms planning budget, while \textit{no} monolithic configuration does; even the fastest monolithic point already exceeds 119\,ms.
This means the planner can dynamically select step count based on the remaining time in each 100\,ms ROS\,2 cycle: when compute is abundant, use $N{=}10$ for maximum accuracy; when a previous module overruns, fall back to $N{=}3$ and still meet the deadline with acceptable quality.

We validate the system in AWSIM closed-loop simulation, running the full Autoware stack with our node as a drop-in replacement for the original monolithic planner~\cite{tieriv_diffusion_planner}.
Over 600+ planning cycles on the Nishishinjuku urban route, the ego vehicle completes the route without collisions, maintaining a consistent $\sim$10\,Hz output rate at 53\,ms inference plus preprocessing.

\section{Discussion and Conclusion}
\label{sec:conclusion}

We presented an open-source modular benchmark that decomposes a monolithic ONNX diffusion planner into independently executable modules within the Autoware production AD stack and AWSIM closed-loop simulator.
The modular design \textit{decouples} architectural and algorithmic optimization: encoder caching alone yields a $3.2\times$ latency reduction, while upgrading to the second-order solver reduces FDE by 41\% at $N{=}3$.
These two improvements are orthogonal and compound while maintaining numerical equivalence.

The decomposition relies only on a repeated substructure conditioned on a cacheable context, a pattern shared by recent diffusion planners~\cite{Zheng2025-eq, Liao2024-js, Liu2025-uh, Zhang2025-eg} and scene generators~\cite{Anguelov2024-rj}, making the approach readily transferable.
To our knowledge, no existing open-source framework provides pluggable solver evaluation under true closed-loop conditions in a ROS\,2-based AD stack.

Our current evaluation covers a single urban scenario in AWSIM.
Broader scenario coverage, systematic GPU latency profiling, and analysis of inter-module transfer overhead at very low step counts are left to future work.
The complete codebase will be released as open-source to support reproducible research in closed-loop diffusion planning.

\bibliographystyle{IEEEtran}
\bibliography{references}

@INPROCEEDINGS{Zheng2025-eq,
  title     = "{Diffusion-based planning for autonomous driving with flexible
               guidance}",
  author    = "Zheng, Yinan and Liang, Ruiming and Zheng, Kexin and Zheng,
               Jinliang and Mao, Liyuan and Li, Jianxiong and Gu, Weihao and Ai,
               Rui and Li, Shengbo Eben and Zhan, Xianyuan and Liu, Jingjing",
  booktitle = "{The Thirteenth International Conference on Learning
               Representations}",
  month     =  "26~" # jan,
  year      =  2025
}

@INPROCEEDINGS{Liao2024-js,
  title     = "{DiffusionDrive: Truncated diffusion model for end-to-end
               autonomous driving}",
  author    = "Liao, Bencheng and Chen, Shaoyu and Yin, Haoran and Jiang, Bo and
               Wang, Cheng and Yan, Sixu and Zhang, Xinbang and Li, Xiangyu and
               Zhang, Ying and Zhang, Qian and Wang, Xinggang",
  booktitle = "{Proceedings of the IEEE/CVF Conference on Computer Vision and Pattern Recognition (CVPR)}",
  publisher = "IEEE",
  month     =  "22~" # nov,
  year      =  2025,
  url       = "https://openaccess.thecvf.com/content/CVPR2025/papers/Liao_DiffusionDrive_Truncated_Diffusion_Model_for_End-to-End_Autonomous_Driving_CVPR_2025_paper.pdf"
}

@INPROCEEDINGS{Liu2025-uh,
  title     = "{BridgeDrive: Diffusion Bridge Policy for Closed-Loop Trajectory
               Planning in Autonomous Driving}",
  author    = "Liu, Shu and Chen, Wenlin and Li, Weihao and Wang, Zheng and
               Yang, Lijin and Huang, Jianing and Zhang, Yipin and Huang,
               Zhongzhan and Cheng, Ze and Yang, Hao",
  booktitle = "{The Fourteenth International Conference on Learning
               Representations}",
  month     =  "8~" # oct,
  year      =  2026,
  url       = "https://openreview.net/forum?id=dJKhjK4zpp"
}

@ARTICLE{Zhang2025-eg,
  title         = "{LAP: Fast LAtent diffusion planner with fine-grained feature
                   distillation for autonomous driving}",
  author        = "Zhang, Jinhao and Xia, Wenlong and Zhou, Zhexuan and Gong,
                   Youmin and Mei, Jie",
  journal       = "arXiv [cs.RO]",
  month         =  "2~" # dec,
  year          =  2025,
  url           = "http://arxiv.org/abs/2512.00470",
  archivePrefix = "arXiv",
  primaryClass  = "cs.RO",
  eprint        = "2512.00470",
  doi           = "10.48550/arXiv.2512.00470"
}

@ARTICLE{Yang2024-dl,
  title         = "{Diffusion-ES: Gradient-free planning with diffusion for
                   autonomous driving and zero-shot instruction following}",
  author        = "Yang, Brian and Su, Huangyuan and Gkanatsios, Nikolaos and
                   Ke, Tsung-Wei and Jain, Ayush and Schneider, Jeff and
                   Fragkiadaki, Katerina",
  journal       = "arXiv [cs.LG]",
  month         =  "9~" # feb,
  year          =  2024,
  url           = "http://arxiv.org/abs/2402.06559",
  archivePrefix = "arXiv",
  primaryClass  = "cs.LG",
  eprint        = "2402.06559",
  doi           = "10.48550/arXiv.2402.06559"
}

@ARTICLE{Xia2025-hx,
  title         = "{DriveLaW:Unifying planning and video generation in a latent
                   driving world}",
  author        = "Xia, Tianze and Li, Yongkang and Zhou, Lijun and Yao,
                   Jingfeng and Xiong, Kaixin and Sun, Haiyang and Wang, Bing
                   and Ma, Kun and Chen, Guang and Ye, Hangjun and Liu, Wenyu
                   and Wang, Xinggang",
  journal       = "arXiv [cs.CV]",
  month         =  "31~" # dec,
  year          =  2025,
  url           = "http://arxiv.org/abs/2512.23421",
  archivePrefix = "arXiv",
  primaryClass  = "cs.CV",
  eprint        = "2512.23421",
  doi           = "10.48550/arXiv.2512.23421"
}

@ARTICLE{Lu2022-nn,
  title         = "{DPM-solver++: Fast solver for guided sampling of diffusion
                   probabilistic models}",
  author        = "Lu, Cheng and Zhou, Yuhao and Bao, Fan and Chen, Jianfei and
                   Li, Chongxuan and Zhu, Jun",
  journal       = "arXiv [cs.LG]",
  month         =  "2~" # nov,
  year          =  2022,
  url           = "http://arxiv.org/abs/2211.01095",
  archivePrefix = "arXiv",
  primaryClass  = "cs.LG",
  eprint        = "2211.01095"
}

@INPROCEEDINGS{Song2020-ip,
  title     = "{Denoising Diffusion Implicit Models}",
  author    = "Song, Jiaming and Meng, Chenlin and Ermon, Stefano",
  booktitle = "{International Conference on Learning Representations}",
  month     =  "2~" # oct,
  year      =  2021,
  url       = "https://openreview.net/forum?id=St1giarCHLP"
}

@INPROCEEDINGS{Ho2020-xt,
  title     = "{Denoising Diffusion Probabilistic Models}",
  author    = "Ho, Jonathan and Jain, Ajay and Abbeel, Pieter",
  booktitle = "{Advances in Neural Information Processing Systems}",
  volume    =  33,
  pages     = "6840--6851",
  year      =  2020,
  url       = "https://proceedings.neurips.cc/paper/2020/hash/4c5bcfec8584af0d967f1ab10179ca4b-Abstract.html"
}

@MISC{onnxruntime2019,
  title        = "{ONNX Runtime: cross-platform, high performance ML inferencing and training accelerator}",
  author       = "{Microsoft}",
  howpublished = "\url{https://onnxruntime.ai/}",
  year         = 2019
}

@MISC{nvidia_graphsurgeon,
  title        = "{ONNX GraphSurgeon}",
  author       = "{NVIDIA}",
  howpublished = "\url{https://github.com/NVIDIA/TensorRT/tree/main/tools/onnx-graphsurgeon}",
  year         = 2024
}

@MISC{tensorrt2024,
  title        = "{NVIDIA TensorRT}",
  author       = "{NVIDIA}",
  howpublished = "\url{https://developer.nvidia.com/tensorrt}",
  year         = 2024
}

@ARTICLE{Guo2024-zx,
  title     = "{EASTER: Learning to split transformers at the edge robustly}",
  author    = "Guo, Xiaotian and Jiang, Quan and Shen, Yixian and Pimentel, Andy
               D and Stefanov, Todor",
  journal   = "IEEE Trans. Comput.-aided Des. Integr. Circuits Syst.",
  publisher = "Institute of Electrical and Electronics Engineers (IEEE)",
  volume    =  43,
  number    =  11,
  pages     = "3626--3637",
  month     =  nov,
  year      =  2024,
  url       = "http://dx.doi.org/10.1109/TCAD.2024.3438995",
  doi       = "10.1109/tcad.2024.3438995",
  issn      = "0278-0070,1937-4151",
  language  = "en"
}

@INPROCEEDINGS{Nair2023-ew,
  title     = "{SecureFrameNet:A rich computing capable secure framework for
               deploying neural network models on edge protected system}",
  author    = "Nair, Renju C and Rao, Madhav and {Muralidhara}",
  booktitle = "{The Third International Conference on Artificial Intelligence
               and Machine Learning Systems}",
  publisher = "ACM",
  address   = "New York, NY, USA",
  pages     = "1--13",
  month     =  "25~" # oct,
  year      =  2023,
  url       = "http://dx.doi.org/10.1145/3639856.3639864",
  doi       = "10.1145/3639856.3639864",
  language  = "en"
}

@INPROCEEDINGS{Jajal2024-rt,
  title     = "{Interoperability in deep learning: A user survey and failure
               analysis of ONNX model converters}",
  author    = "Jajal, Purvish and Jiang, Wenxin and Tewari, Arav and Kocinare,
               Erik and Woo, Joseph and Sarraf, Anusha and Lu, Yung-Hsiang and
               Thiruvathukal, George K and Davis, James C",
  booktitle = "{Proceedings of the 33rd ACM SIGSOFT International Symposium on
               Software Testing and Analysis}",
  publisher = "ACM",
  address   = "New York, NY, USA",
  volume    =  18,
  pages     = "1466--1478",
  month     =  "11~" # sep,
  year      =  2024,
  url       = "http://dx.doi.org/10.1145/3650212.3680374",
  doi       = "10.1145/3650212.3680374",
  language  = "en"
}

@MISC{apollo2024,
  title        = "{Apollo: Open source autonomous driving platform}",
  author       = "{Baidu}",
  howpublished = "\url{https://github.com/ApolloAuto/apollo}",
  year         = 2024
}

@MISC{awsim2024,
  title        = "{AWSIM: Open-source autonomous driving simulator for Autoware}",
  author       = "{TIER IV, Inc.}",
  howpublished = "\url{https://github.com/tier4/AWSIM}",
  year         = 2024
}

@MISC{tieriv_diffusion_planner,
  title        = "{Autoware Diffusion Planner}",
  author       = "{TIER IV, Inc.}",
  howpublished = "\url{https://github.com/autowarefoundation/autoware_universe/tree/main/planning/autoware_diffusion_planner}",
  year         = 2025
}

@TECHREPORT{autoware2025e2e,
  title       = "{Autoware: Introducing Open-Source End-to-End Autonomous Driving}",
  author      = "{The Autoware Foundation}",
  institution = "{The Autoware Foundation}",
  year        = 2025,
  note        = "Technical Report. Available at \url{https://autoware.org/wp-content/uploads/2025/10/Autoware-E2E-Report_Release.pdf}"
}

@INPROCEEDINGS{Karnchanachari2024-ko,
  title     = "{Towards learning-based planning: The nuPlan benchmark for
               real-world autonomous driving}",
  author    = "Karnchanachari, Napat and Geromichalos, Dimitris and Tan, Kok
               Seang and Li, Nanxiang and Eriksen, Christopher and Yaghoubi,
               Shakiba and Mehdipour, Noushin and Bernasconi, Gianmarco and
               Fong, Whye Kit and Guo, Yiluan and Caesar, Holger",
  booktitle = "{2024 IEEE International Conference on Robotics and Automation
               (ICRA)}",
  publisher = "IEEE",
  volume    =  56,
  pages     = "629--636",
  month     =  "13~" # may,
  year      =  2024,
  url       = "http://dx.doi.org/10.1109/ICRA57147.2024.10610077",
  doi       = "10.1109/icra57147.2024.10610077",
  language  = "en"
}

@INPROCEEDINGS{Dauner2023-eg,
  title     = "{Parting with Misconceptions about Learning-based Vehicle Motion
               Planning}",
  author    = "Dauner, Daniel and Hallgarten, Marcel and Geiger, Andreas and
               Chitta, Kashyap",
  booktitle = "{Conference on Robot Learning}",
  publisher = "PMLR",
  pages     = "1268--1281",
  month     =  "2~" # dec,
  year      =  2023,
  url       = "https://proceedings.mlr.press/v229/dauner23a.html",
  issn      = "2640-3498",
  language  = "en"
}

@INPROCEEDINGS{Peebles2023-vq,
  title     = "{Scalable diffusion models with transformers}",
  author    = "Peebles, William and Xie, Saining",
  booktitle = "{Proceedings of the IEEE/CVF international conference on computer
               vision}",
  pages     = "4195--4205",
  year      =  2023,
  url       = "http://openaccess.thecvf.com/content/ICCV2023/html/Peebles_Scalable_Diffusion_Models_with_Transformers_ICCV_2023_paper.html"
}

@INPROCEEDINGS{Song2020-dn,
  title     = "{Score-Based Generative Modeling through Stochastic Differential
               Equations}",
  author    = "Song, Yang and Sohl-Dickstein, Jascha and Kingma, Diederik P and
               Kumar, Abhishek and Ermon, Stefano and Poole, Ben",
  booktitle = "{International Conference on Learning Representations}",
  month     =  "2~" # oct,
  year      =  2020,
  url       = "https://openreview.net/forum?id=PxTIG12RRHS&utm_campaign=NLP%20News&utm_medium=email&utm_source=Revue%20newsletter"
}

@INPROCEEDINGS{Betz2023-zb,
  title     = "{How fast is my software? Latency evaluation for a ROS 2
               autonomous driving software}",
  author    = "Betz, Tobias and Schmeller, Maximilian and Teper, Harun and Betz,
               Johannes",
  booktitle = "{2023 IEEE Intelligent Vehicles Symposium (IV)}",
  publisher = "IEEE",
  pages     = "1--6",
  month     =  "4~" # jun,
  year      =  2023,
  url       = "http://dx.doi.org/10.1109/IV55152.2023.10186585",
  doi       = "10.1109/iv55152.2023.10186585",
  issn      = "2642-7214,1931-0587",
  language  = "en"
}

@ARTICLE{Betz2025-at,
  title     = "{End-to-end latency optimization for containerized ROS 2
               autonomous driving software}",
  author    = "Betz, Tobias and Teper, Harun and Ebner, Dominic and Leitenstern,
               Maximilian and Sagmeister, Simon and Weinmann, Marcel and Chen,
               Jian-Jia and Lienkamp, Markus",
  journal   = "IEEE Access",
  publisher = "Institute of Electrical and Electronics Engineers (IEEE)",
  volume    =  13,
  pages     = "112654--112672",
  year      =  2025,
  url       = "http://dx.doi.org/10.1109/ACCESS.2025.3582868",
  doi       = "10.1109/access.2025.3582868",
  issn      = "2169-3536"
}

@INPROCEEDINGS{Likhachev2003-tx,
  title     = "{ARA*: Anytime A* with provable bounds on sub-optimality}",
  author    = "Likhachev, Maxim and Gordon, Geoffrey J and Thrun, S",
  booktitle = "{Advances in Neural Information Processing Systems}",
  volume    =  16,
  pages     = "767--774",
  month     =  "9~" # dec,
  year      =  2003,
  url       = "https://proceedings.neurips.cc/paper/2003/hash/ee8fe9093fbbb687bef15a38facc44d2-Abstract.html"
}

@INPROCEEDINGS{Janner2022-gh,
  title     = "{Planning with Diffusion for Flexible Behavior Synthesis}",
  author    = "Janner, Michael and Du, Yilun and Tenenbaum, Joshua and Levine,
               Sergey",
  booktitle = "{International Conference on Machine Learning}",
  publisher = "PMLR",
  pages     = "9902--9915",
  month     =  "28~" # jun,
  year      =  2022,
  url       = "https://proceedings.mlr.press/v162/janner22a.html",
  issn      = "2640-3498",
  language  = "en"
}

@INPROCEEDINGS{Chi2023-bj,
  title     = "{Diffusion policy: Visuomotor policy learning via action
               diffusion}",
  author    = "Chi, Cheng and Feng, Siyuan and Du, Yilun and Xu, Zhenjia and
               Cousineau, Eric and Burchfiel, Benjamin and Song, Shuran",
  booktitle = "{Robotics: Science and Systems XIX}",
  publisher = "Robotics: Science and Systems Foundation",
  month     =  "10~" # jul,
  year      =  2023,
  url       = "http://dx.doi.org/10.15607/rss.2023.xix.026",
  doi       = "10.15607/rss.2023.xix.026"
}

@INPROCEEDINGS{Anguelov2024-rj,
  title     = "{SceneDiffuser: Efficient and controllable driving simulation
               initialization and rollout}",
  author    = "Jiang, Chiyu Max and Bai, Yijing and Cornman, Andre and Davis,
               Christopher and Huang, Xiukun and Jeon, Hong and Kulshrestha,
               Sakshum and Lambert, John and Li, Shuangyu and Zhou, Xuanyu and
               Fuertes, Carlos and Yuan, Chang and Tan, Mingxing and Zhou, Yin
               and Anguelov, Dragomir",
  editor    = "Globerson, A and Mackey, L and Belgrave, D and Fan, A and Paquet,
               U and Tomczak, J and Zhang, C",
  booktitle = "{Advances in Neural Information Processing Systems 37}",
  publisher = "Neural Information Processing Systems Foundation, Inc. (NeurIPS)",
  address   = "San Diego, California, USA",
  volume    =  37,
  pages     = "55729--55760",
  year      =  2024,
  url       = "http://dx.doi.org/10.52202/079017-1771",
  doi       = "10.52202/079017-1771"
}

\end{document}